\pdfoutput=1

\documentclass[11pt]{article}

\usepackage[final]{acl}

\usepackage{times}
\usepackage{latexsym}
\usepackage{setspace}

\usepackage{graphicx}
\usepackage{tabularx}
\usepackage{multirow} 
\usepackage{soul}
\usepackage{xstring}
\usepackage{caption}
\usepackage{tikz}

\usepackage{url}
\usepackage[most]{tcolorbox}

\usepackage[T1]{fontenc}

\usepackage[utf8]{inputenc}

\usepackage{microtype}

\usepackage{inconsolata}

\usepackage{color}

\usepackage{setspace}
\title{Language Models for Text Classification: \\ Is In-Context Learning Enough?}

\author{Aleksandra Edwards \and Jose Camacho-Collados \\
  \texttt{edwardsai, camachocolladosj@cardiff.ac.uk} \\}

\begin{document}
\maketitle

\begin{spacing}{0.99}

\begin{abstract}
Recent foundational language models have shown state-of-the-art performance in many NLP tasks in zero- and few-shot settings. An advantage of these models over more standard approaches based on fine-tuning is the ability to understand instructions written in natural language (prompts), which helps them generalise better to different tasks and domains without the need for specific training data. This makes them suitable for addressing text classification problems for domains with limited amounts of annotated instances. However, existing research is limited in scale and lacks understanding of how text generation models combined with prompting techniques compare to more established methods for text classification such as fine-tuning masked language models. In this paper, we address this research gap by performing a large-scale evaluation study for 16 text classification datasets covering binary, multiclass, and multilabel problems. In particular, we compare zero- and few-shot approaches of large language models to fine-tuning smaller language models. We also analyse the results by prompt, classification type, domain, and number of labels. In general, the results show how fine-tuning smaller and more efficient language models can still outperform few-shot approaches of larger language models, which have room for improvement when it comes to text classification. 

\end{abstract}

\section{Introduction}\label{intro}

A standard approach for supervised text classification is fine-tuning language models such as BERT using an additional classifier head~\cite{radford2018improving, dong2019unified, devlin2018bert, yin2019benchmarking, viswanathan2023prompt2model, mosbach2023few}. However, these approaches require large amounts of data to achieve state-of-the-art results~\cite{edwards2022guiding} which makes them unsuitable for classification tasks associated with class imbalances and data sparsity~\cite{giridhara2019study, zhang2015short, turker2019knowledge, yin2019benchmarking}. These  problems often occur in real world applications where annotation of data can be performed only by scarce domain experts such as medical and legal domains or applications with highly imbalanced classes such as crime data and fraud detection~\cite{giridhara2019study, zhang2015short, turker2019knowledge}.

Recent advances in Natural Language Processing (NLP) lead to the emerge of an alternative approach based on using autoregressive text generation models~\cite{radford2019language} that have zero- and few- shot capabilities and perform unseen tasks through the use of 
prompting~\cite{schick2021exploiting, radford2019language, le2021many, viswanathan2023prompt2model, plaza2023leveraging}. The ability of these models to understand natural language instructions let them generalise to different domains and tasks without the need of large training corpora~\cite{plaza2023leveraging}. There have been even further improvements in the performance of these models in zero-shot settings by fine-tuning them on sets of instructions (task descriptions)~\cite{raffel2020exploring}.

The promising results of these models against various benchmark datasets~\cite{wang2022super, liu2023gpteval, bang2023multitask} led to increased research into developing methods, mainly based on prompt engineering techniques~\cite{viswanathan2023prompt2model, le2021many} for improving their generalisation capabilities. Further, there has been an increased attention into evaluating the suitability of these models for more specialised domains such as the legal, medical, and financial domain~\cite{sarkar-etal-2021-shot, chalkidis2020empirical, yin2019benchmarking,labrak2023zero}. However, most of the proposed approaches are domain- and task-specific. There is lack of understanding of how these models perform in comparison to more established approaches for text classification. In general, analyses are performed for a small range of model types, domains, and tasks.

Our work is the first attempt to systematically compare how text generation models using zero-shot and one-shot learning compare to more established but data-consuming approaches for classification based on fine-tuning language models. Our goal is to identify how well current large language models (LLMs) can adapt to different text classification tasks and domains given limited information, and outline the potential strengths and weaknesses of these models. 
For these purposes, we evaluate five heterogeneous models of different sizes, including traditional masked language models and more recent autoregressive LLMs.  
Our analyses span over 16 datasets from 7 domains representing binary, multiclass, and multilabel classification. 

Our main contributions are as follows. First, we explore an important but understudied problem of how suitable the newly developed text generation models such as LLaMA, Flan-T5, T5, and ChatGPT are for text classification in few-shot settings compared to lighter models that require training data such as RoBERTa or FastText. 
In addition to the performance, our analysis helps identify specific strengths and weaknesses of each type of model. 
Second, in contrast to the majority of existing research focusing on optimisation techniques for prompt creation, we analyse trends in the model's performance that are non-prompt sensitive as well as look at how the amount of specificity provided in the prompt regarding the task and the domain affect the performance of the models. Third, we evaluate generalisation abilities of models for 7 domains, including real-world specialised domains, such as legal, medical, and crime data. We also analyse how the models' behaviour changes for datasets used in the pre-training stage versus when testing on unseen datasets. 

\section{Related Work}\label{relwork}
    We first introduce the different types of methods and models used for text classification along with their strengths and weaknesses (see Section \ref{textclassification}). Then, we discuss relevant work on comparing prompting and fine-tuning approaches for text classification as well as outline challenges and research gaps within existing work (see Section \ref{finetuning_vs_prompting}). 

    \subsection{Text Classification}\label{textclassification}
    We distinguish between three main approaches for text classification, linear methods (described in Section \ref{traditional}), fine-tuning language models (Section \ref{finetuning}) and prompting techniques combined with text generation models (Section \ref{textgenerationmodels}).
    \subsubsection{Linear Methods}\label{traditional}
        FastText \cite{joulin2017bag} is a linear text classification model which provides a strong baseline for many text classification tasks and gives performance comparable to state-of-the-art methods, including language models such as BERT \cite{10.1145/3397056.3397082, edwards2020go}. It integrates a linear model with a rank constraint which allows sharing parameters among classes and features. It also integrates word embeddings which are averaged into a text. These features help address many problems associated with other linear models such as out-of-vocabulary words and fine-grained distinctions between classes.  
        \subsubsection{Fine-tuning Methods}\label{finetuning}
     Language models like BERT~\cite{devlin2018bert} and RoBERTa~\cite{liu2019roberta}, pre-trained using a masked language modeling (MLM) objective provide a state-of-the-art performance against most standard NLP benchmarks~\cite{wang2019superglue, wang2018glue}. These models can be easily adapted for text classification by using fine-tuning techniques which are based on adding a single classification layer onto the model.  However, fine-tuning techniques require large amounts of data to be adapted to targeted tasks and domains which makes them impractical for low resource classification tasks~\cite{strubell2019energy, peng2021domain, lu2021sentence}. 
     
    \subsubsection{Text Generation Models}\label{textgenerationmodels}
    Recent advances in NLP have led to the development of bigger models composed of billion of parameters which have shown an improved performance especially in text generation and low resource settings~\cite{zhang2019bertscore, black2022gpt, labrak2023zero}. These text generation models such as GPT and subsequent releases~\cite{brown2020language, radford2018improving, radford2019language} as well as LLaMA~\cite{touvron2023llama, touvron2023llama2} and T5~\cite{raffel2020exploring} can understand natural language instructions (i.e., prompts) and thus can generalise to unseen tasks and domains without the need for large computational and data resources~\cite{brown2020language}. Further progress has been made by fine-tuning these models on a set of natural language instructions, consisting of descriptions of the tasks and the expected output~\cite{efrat2020turking, mishra2022cross}. This enables models to generalise even better to tasks, domains, and languages ~\cite{ouyang2022training, wei2021finetuned, sanh2022multitask, efrat2020turking, mishra2022cross}.

    The ability of text generation models to make predictions with little or no training makes these models particularly suitable for tackling the problem of data scarcity for text classification~\cite{wang2020generalizing, gupta2020effective}.  Therefore, much of the approaches in zero- and few-shot learning  are focused at optimising the performance of these models mainly through the use of prompting~\cite{gera-etal-2022-zero, le2021many, deng2022rlprompt, schick2021exploiting, radford2019language, le2021many, viswanathan2023prompt2model, plaza2023leveraging}.

    \subsection{Prompting versus Fine-Tuning}\label{finetuning_vs_prompting}
     Prompting in zero- and few- shot settings, also known as in-context learning (ICL), is the process of providing natural language instructions that describe a task as an input to a language model, including the expected output~\cite{labrak2023zero}. In few-shot prompting, the model is presented with some training examples along with the task instructions. In contrast to fine-tuning techniques, prompting does not involve changing the weights of the model which makes the approach less resource consuming. Additionally, previous research has suggested that prompting can lead to comparable or even better performance than standard fine-tuning techniques~\cite{gao2021making, mosbach2023few}. A drawback of this approach is the models' sensitivity to the prompts where slight changes of the instruction can lead to big differences in the performance~\cite{schick2021s,le2021many,sun2023evaluating}. Thus, much of the work on text generation models is focused on prompt optimisation techniques based on automatic generation for prompts~\cite{wang2022automatic,shin2020autoprompt}, quantifying the benefits of prompting~\cite{schick2021s,le2021many}, and improving the generalisation abilities of prompts~\cite{zhang-etal-2022-learn, schonfeld2019generalized, song2021generalized,wang2022automatic,oniani2023large,sun2023evaluating}

    There has been an increased research into evaluating and improving the performance of text generation models for zero and few shot classification in more specialised domains such as the legal, medical, and financial domains~\cite{ge2022few,sarkar-etal-2021-shot, chalkidis2020empirical}. \newcite{labrak2023zero} evaluate four state-of-the-art instruction-tuned large language models (ChatGPT, Flan-T5 UL2, Tk-Instruct, and Alpaca) on a set of 13 real-world clinical and biomedical natural NLP tasks, including text classification. The results show that instruction-tuned models tend to be outperformed by a specialised model trained for the medical field such as PubMedBERT~\cite{gu2021domain}. This rises questions into the suitability of text generation models and prompting techniques for more specialised domains which require domain experts for annotation. Another research by \newcite{mosbach2023few} conducts a comparison between fine-tuning and prompting techniques for two text classification datasets showing that both approaches have similar performance, although with a large variation in results depending on properties such as model size and number of examples. These works show that adapting these models to tasks, especially text classification for more specialised domains, remains a challenge. 

    The variance in performance between tasks and models depending on the prompt design makes the generalisation of text generation models a challenging problem. The small scale on which analyses are performed does not give enough knowledge on how well prompting techniques compare to the more established models for classification across different text classification types and more challenging unfamiliar domains. In this paper, we address these challenges by performing a large-scale comparison between different model types across a wider range of classification tasks and domains. 

    \section{Experimental Setting}
    \subsection{Datasets}\label{datasets}
     \begin{table*}[hbt!]
	\centering
     \large
	
	\resizebox{\linewidth}{!}{
	
		\begin{tabular}{|l|l|l|l|l|c|c||c|c|c|}\hline
		  \textbf{Dataset}&\textbf{Domain}&\textbf{Task}&\textbf{Type}&\textbf{Class Type}&\textbf{Avg tokens}&\textbf{Labels}&\textbf{\# Train}&\textbf{\# Dev}&\textbf{\# Test}\\\hline
		  SemEval 18 (Emoji)&Twitter&Emoji Prediction&Sentence&multiclass&12&20&45,000&5,000&50,000\\\hline
		  SemEval 18 (Irony)&Twitter&Irony Detection&Sentence&binary&13&2&2,862&955&784\\\hline
		  SemEval 19 (Hateval)&Twitter&Hateval&Sentence&binary&18&2&9,000&1,000&2,970\\\hline
		 SemEval 19 (OffensEval)&Twitter&OffensEval&Sentence&binary&19&2&11,916&1,324&860\\\hline
		 SemEval 17 (Sentiment)&Twitter&Sentiment Analysis&Sentence&multiclass&20&3&45,389&2,000&11,906\\\hline
		  BBC news&News&Topic categorisation&Document&multiclass&220&5&1602&178&445\\\hline
		  Reuters&News&Topic categorisation&Document&multiclass&83&8&6120&680&2659\\\hline
		  AG News&News&Topic categorisation&Document&multiclass&31&4&103,346&11,482&5,928\\\hline
		  20 Newsgroups&News&Topic categorisation&Document&multiclass&285&6&9,857&1,095&7,290\\\hline
		   20 Newsgroups&News&Topic categorisation&Document&multiclass&285&20&9,857&1,095&7,290\\\hline
		 IMDB reviews&Reviews&Polarity Detection&Document&binary&231&2&25200&2,800&25,601\\\hline
        Ohsumed&Medical&Cardiovascular diesese det.&Document&multiclass&104&23&9,390&1,043&12733\\\hline
		  Toxic Comments&Wikipedia&Toxic prediction&Document&multilabel&46&7&143,614&15,957&63,978\\\hline
		  PCL dataset&News&Patronising language det.&Document&multilabel&37&7&517&57&419\\\hline
		  EU legislation documents&Legislation&Legal legislation concept det.&Document&multilabel&27&10&45,000&6,000&6,000\\\hline
		  Hallmarks of cancer&Medical&Hallmarks of cancer detection&Sentence&multilabel&22&10&12,456&1,384&3624\\\hline
		  Safeguarding reports&Safeguarding&Theme detection&Sentence&multilabel&18&5&5,719&635&3496\\\hline
		 Safeguarding reports&Safeguarding&Theme detection&Sentence&multilabel&18&10&5,719&635&3496\\\hline
		\end{tabular}
		}
		\caption{Overview of the classification datasets used in our experiments.}\label{tab:desc}
		    \end{table*}
    For our experiments we selected a suite of datasets representing all three classification types, i.e., binary, multiclass, and multilabel.
    The datasets span across 7 domains and 13 classification tasks. Specifically, we selected the Twitter datasets from the SemEval 18 on emoji prediction \cite{barbieri2018semeval}, SemEval 18 on irony Detection \cite{van2018semeval}, SemEval 19 on hate detection \cite{basile2019semeval}, SemEval 19 on offense detection \cite{zampieri2019semeval}, and SemEval 19 on sentiment analysis \cite{nakov2019semeval}. Further, we include datasets for topic categorisation such as BBC news\footnote{http://mlg.ucd.ie/datasets/bbc.html}, AG News~\citep{zhang2015character}, Reuters~\citep{lewis2004rcv1}, and 20 Newsgroups~\citep{lang1995newsweeder} , as well as IMDB reviews dataset for polarity detection~\citep{maas2011learning}, PCL dataset for patronising language detection~\citep{perez-almendros-etal-2020-dont}, and Toxic comments~\citep{hosseini1702deceiving}. Additionally, we  evaluate models for more specialised domains representing real world applications such as EU legislation documents~\citep{chalkidis2019large} for legal legislation concepts detection, Hallmarks of cancer~\citep{baker2015automatic} for detecting cancer hallmarks, Ohsumed~\citep{joachims1998text} for cardiovascular diseases detection, and Safeguarding reports~\citep{edwards2022guiding} for theme detection. Additionally, we perform prediction for the top classes as well as the sub-classes of the 20 Newsgroups and Safeguarding datasets. In this way, we can analyse how the models performance is affected by the number of classes. The main features and statistics of each dataset are summarized in Table~\ref{tab:desc}. For the EU legislation documents we have performed experiments with the 10 most frequent labels, similarly to \newcite{chalkidis2019large}. For the Ohsumed dataset, we have selected the top 23 most frequent classes, 
    similar to prior work~\cite{pilehvar2017towards}.

      \subsection{Comparison Models}\label{comparisonmodels}
   
      We compare three main types of models: generative language models, masked language models, and linear models, all described below. 

\noindent \textbf{Generative Language Models.}
     We include LLaMA 1~\cite{touvron2023llama} and  2~\cite{touvron2023llama2} into the analysis as representatives of large auto-regressive generation models, both with 7 billion parameters. As a representative of smaller but instruction-tuned model, we use Flan-T5~\cite{chung2022scaling}. The model is fine-tuned using the Flan instruction tuning tasks collection~\citep{chung2022scaling}. We use the large Flan-T5 model with 780M parameters. We have also included T5 model \cite{raffel2020exploring} into our analysis which we fine-tune, similarly to RoBERTa. In particular, we use T5 base model. We have downloaded the models from Hugging Face~\cite{wolf2019huggingface}. 
     As a representative of the GPT family of autoregressive models \cite{brown2020language}, we use OpenAI GPT 3.5-Turbo for our analysis. We added this model for completeness. However, given budget constraints and its closed nature for which few conclusions can be drawn, we only provide results for a sample of all datasets.      
     
\noindent \textbf{Masked Language Models.} As a representative of masked language model, we use RoBERTa~\cite{liu2019roberta}, pre-trained on English language. It is known to achieve state-of-the-art results for many text classification tasks. We perform experiments with RobERTa base (125 million parameters) and RoBERTa large (354 million parameters) models to allow analysis into the effect of model size over the classification performance. We have downloaded the models from Hugging Face~\cite{wolf2019huggingface}.

   \noindent \textbf{Linear Models.} Finally, we use FastText \cite{joulin2017bag} (see Section \ref{traditional}) as a representative of a linear text classification model. Despite its simplicity the model provides a strong baseline for many text classification tasks and it is known to give comparable results to state-of-the-art methods, including language models such as BERT for some classification problems \cite{10.1145/3397056.3397082, edwards2020go}.

    \subsection{Prompting, Training and Evaluation}\label{training}
    As mentioned in Section~\ref{intro}, our aim is to estimate how well the text generation models perform for text classification when 
    compared to the more data consuming models such as RoBERTa and FastText. Therefore, we perform experiments for Flan-T5 and LLaMA in zero- and one- shot ICL settings. For zero shot, we provide information about the task to the model through prompting. For one shot, we randomly select a single training instance per label and we provide these examples along with the instruction to the model. To ensure robustness, the random selection of training samples is performed for three iterations and the results are averaged. For generating labels for the test sequences, we use default model settings. We judge the outputs as expected class labels or not by simply checking whether the output of the model matches one of the labels for the given classification task. We experiment with three different prompts which we describe further in Section~\ref{prompts}. 
    
    As for RoBERTa, we fine-tune it for the classification task on the training data of each dataset using a sequence classifier, a learning rate of 2e-5 and 4 epochs. In particular, we made use of RoBERTa’s Hugging Face default transformers implementation for classifying sentences~\cite{wolf2019huggingface}. As for T5, we fine-tune it using conditional generation, 2 epochs, and learning rate of 5e-5. Finally, we use FastText classifier with 25 epochs and softmax as the loss function. 
    
    Finally, we report results based on standard micro and macro averaged F1~\citep{yang1999evaluation}.
    
    \subsection{Prompt Design}\label{prompts}

    Our paper does not focus on identifying and describing most efficient prompt engineering practices (as majority of work described in Section~\ref{relwork}) but instead we focus on highlighting prompt-independent trends in the models performance in order to help outline advantages and disadvantages of out-of-the-box approaches for few shot text classification. We selected instructions that led to satisfactory results in previous research or have been used in the training set for the instruction-tuned models Flan-T5~\cite{sun2023evaluating, wei2021finetuned}. These prompts vary in the detail they provide about the given task and domain. We want to analyse trends across models behaviour that are non-prompt sensitive as well as look at how the amount of specificity provided in the prompt affect the performance of the models. For these purposes, we use the following three prompts: (1) \textit{\textbf{generic:}} a prompt which does not give information about the task or domain, used in \cite{sun2023evaluating}; (2) \textit{\textbf{task:}} describes the given task, i.e., classification; (3) \textit{\textbf{domain:}} a prompt which gives more information about the domain, for instance, it specifies the type of test data, such as an article or tweet. We have created the domain-based prompts following examples provided in  \newcite{wei2021finetuned}. Table~\ref{tab:prompts} presents examples of the prompts per classification type\footnote{A list of all prompts is given in the Appendix.}.

    \begin{table}[hbt!]
    \LARGE
	\centering
	\resizebox{\linewidth}{!}{
	\begin{tabular}{|l|p{4.7cm}|p{5.3cm}|p{6.0cm}|}\hline
		&\textbf{Binary}&\textbf{Multiclass}&\textbf{Multilabel}\\\hline
		\textbf{generic}&Choose your answer: According to the above paragraph, the question \textit{'Is the text ironic?'}:&Pick one category for the following text. The options are:&Pick one or more from the categories for the following text.The options are:\\\hline
		\textbf{task}&Classify the input text into one of the following categories:&Classify the input text into one of the following categories:&Classify the input text using one or more from the following categories:\\\hline
		\textbf{domain}&Is the Tweet classified as irony or non-irony?&Select the topic that the given news is about. The topics are - & Which of the given toxic topics best describe the given comment? Choose one or more from the following topics:\\\hline

	\end{tabular}
	}
	\caption{Examples of prompts used for zero- and one-shot learning for Flan-T5 and LLaMA.}\label{tab:prompts}
	\end{table}

 \section{Results and Analysis}\label{results}
  
    The aim of our analysis is (1) identify if and how the use of prompts affect the performance of text generation models (see Section \ref{promptinganalysis}); (2) compare performance of prompting and fine-tuning techniques in order to identify strengths and weaknesses of the different models -- we focus on a comparison between the three types of classification, i.e., binary, multiclass, and multilabel (see Section \ref{promptvsfinetune}); and (3) perform a fine-grained analysis comparing models' performance at the domain and dataset level (see Section ~\ref{comparison_bestperforming}). In addition to this general comparison, we analyse separately the performance of closed-source GPT3.5 and models for the `IMDB reviews' and `AG News' datasets as they are used in the fine-tuning of the Flan-T5 model.

 \subsection{Model and Prompt  Analysis}\label{promptinganalysis}

\begin{figure*}[hbt!]
		    \begin{center}
		     \includegraphics[scale = 0.04]{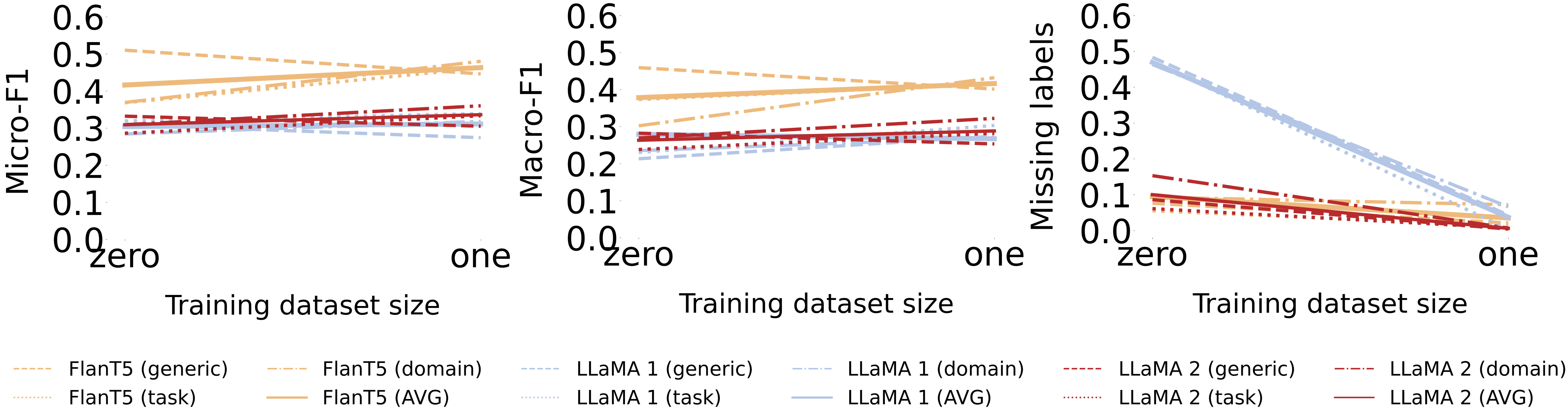}
            \caption{Micro-F1 (left) and Macro-F1 (middle) results averaged across all datasets, comparing the performance of Flan-T5, LLaMA 1, and LLaMA 2 models for all three types of prompts, i.e., \textit{'generic'}, \textit{'task'}, and \textit{'domain'} as well as the average ('AVG') between them. 'Missing label' (right) shows the fraction of results returned by the three models that are different from the classification labels. Results are displayed for zero-shot ('zero') and one-shot setting ('one').}\label{promptscomparison}
		    \end{center}
        \end{figure*}

        A comparison between the two LLaMA models shows an advantage of LLaMA 2 over LLaMA 1 for both zero- and one-shot settings across all prompt types (see Figure~\ref{promptscomparison} and Table~\ref{tab:promptanalysis}). 
        The two models have similar performance in the zero-shot setting in terms of F1 score. However, the number of wrong labels for LLaMA 1 is much larger with 0.470 wrong labels compared to the 0.100 wrong labels from LLaMA2. Results in Figure~\ref{promptscomparison} also show a clear advantage of Flan-T5 over the other models for all three prompts in terms of micro- and macro- F1 for both zero- and one- shot settings. The Flan-T5 model also leads to smaller number of wrong labels in zero-shot prompting. This suggests that smaller but instruction-tuned models can be more beneficial in zero- and few- shot classification in comparison to larger text generation models. Specifically, Flan-T5 has on average 0.110 improvement in micro- and macro-F1 for both zero- and one- shot settings over LLaMA 2. 
        \begin{table*}[!t]

	       \centering
	       \setlength{\tabcolsep}{6.0pt}
	       \scalebox{0.68}{
	
		  \begin{tabular}{|l|l||c|c|c|c|c|c|c|c|}\hline
		      \multirow{2}{*}{\textbf{Model}}&\multirow{2}{*}{\textbf{Prompt}}&\multicolumn{3}{|c|}{\textbf{zero shot}}&\multicolumn{3}{|c|}{\textbf{one shot}}&\multicolumn{2}{|c|}{\textbf{all}}\\\cline{3-10}
		          &&\textbf{micro F1}&\textbf{macro F1}&\textbf{missing labs}&\textbf{micro F1}&\textbf{macro F1}&\textbf{missing labs}&\textbf{micro F1}&\textbf{macro F1}\\\hline\hline

          \multirow{4}{*}{Flan-T5}
        &generic&.510&.459&.076&.446&.401&.020&--&---\\\cline{2-10}
        &task&.368&.373&.055&.462&.415&.012&--&---\\\cline{2-10}
		&domain&.369&.302&.092&.480&.432&.072&--&---\\\cline{2-10}
        &AVG&.416&.378&.074&.463&.416&.035&---&---\\\hline\hline
       
        \multirow{4}{*}{LLaMA 1}
        &generic&.309&.213&.484&.274&.274&.043&--&---\\\cline{2-10}
        &task&.319&.230&.471&.339&.303&.414&--&---\\\cline{2-10}
		&domain&.284&.235&.463&.318&.270&.066&--&---\\\cline{2-10}
        &AVG&.304&.279&.469&.311&.267&.038&--&---\\\hline\hline

        \multirow{4}{*}{LLaMA 2}
        &generic&.332&.282&.086&.305&.253&.436&--&---\\\cline{2-10}
        &task&.286&.238&.061&.333&.282&.679&--&---\\\cline{2-10}
		&domain&.309&.269&.153&.360&.322&.007&--&---\\\cline{2-10}
        &AVG&.309&.263&.100&.336&.288&.006&--&---\\\hline\hline

        T5&---&--&---&---&.134&.109&.851&.702&.625\\\hline\hline
        RoBERTa (base)&---&--&---&---&.273&.207&---&.707&.625\\\hline\hline
        RoBERTa (large)&---&--&---&---&.338&.278&---&.727&.657\\\hline\hline
		fastText&---&---&---&---&.254&.164&---&.505&.419\\\hline\hline

        \end{tabular}
		}
	\caption{Prompt Analysis where Micro-F1 and Macro-F1 results averaged across all datasets, comparing the performance of Flan-T5, LLaMA 1, and LLaMA 2 models for all three types of prompts, i.e., \textit{`generic'}, \textit{`task'}, and \textit{`domain'} as well as the average (`AVG') between them. `Missing labs' shows the fraction of results returned by the three models that are different from the classification labels. Results are displayed for zero-shot (`zero') and one-shot setting (`one').}\label{tab:promptanalysis}
	\end{table*}
 
        Further analysis into the prompts reveal that prompt choice does not lead to significant changes in the models behaviour where the deviation for the three prompts across all models is relatively small. For instance, for LLaMA 1 and LLaMA 2 is less than 0.02 difference in micro-F1 for both zero- and one- shot settings while for Flan-T5 it gradually decreases from 0.07  in zero-shot to 0.01 for one-shot. This suggests that smaller models such as Flan-T5 are more sensitive to the prompt in zero settings versus few shot learning. The benefits from one-shot prompting are evident across all three models where the F1 measure tends to increase and the number of wrong labels decreases. Flan-T5 improves its performance on a higher rate compared to to the other two models with around 0.047 increase in the micro-F1 score versus 0.027 increase for LLaMA 2. This illustrates the strong abilities of these models to learn tasks with minimal amount of training data.

 \subsection{Prompting versus Fine-tuning}\label{promptvsfinetune}
\begin{figure}[hbt!]
		    \begin{center}
		     \includegraphics[scale = 0.02]{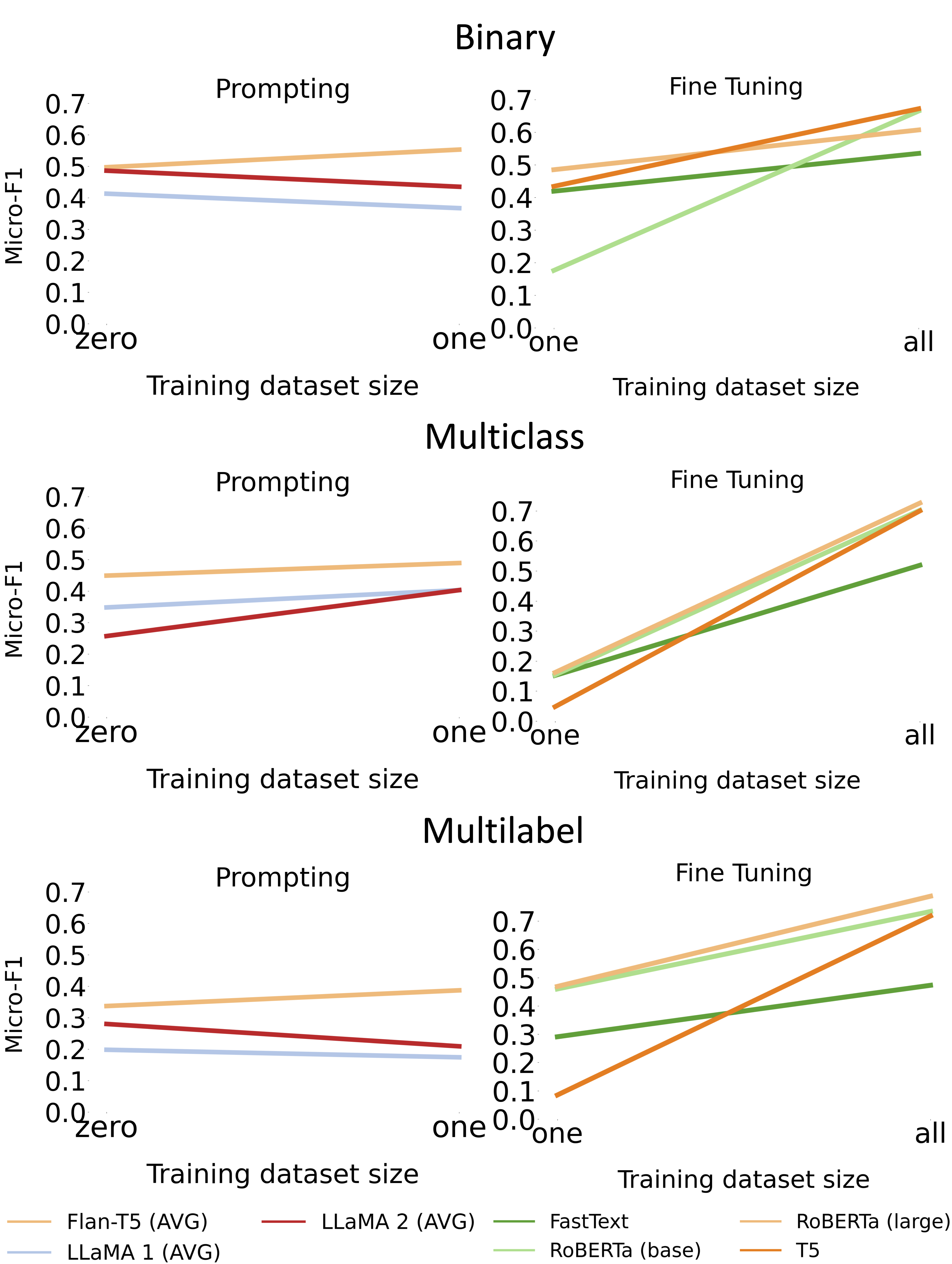}
        \caption{Comparison between prompting (left) and fine-tuning (right) approaches per text classification type  where 'AVG' refers to averaged results across all prompt types per model. In 'Prompting', 'zero' and 'one' refer to zero- and one- shot prompt-based learning techniques, in 'Fine Tuning', 'one' refers to fine-tuning the models with one training instance per label and 'all' refers to fine-tuning using the entire dataset.}\label{classtypeall}
		    \end{center}
        \end{figure}

  Results in Figure~\ref{classtypeall} show the same trends for prompting methods where Flan-T5 outperforms LLaMA 1 and LLaMA 2 for all text classification types in terms of micro- and macro-F1. All three models improve their performance for one-shot prompting regarding the number of wrong labels.  In one shot setting, Flan-T5 and LLaMA 2 tend to have close to 0 wrong labels with LLaMA 2 returning slightly lower number of irrelevant results, while Flan-T5 has a better F1 score (see Figures~\ref{classtypeall} and ~\ref{missinglabels}\footnote{Macro-F1 results are available in the Appendix.}).  The advantage of LLaMA 2 over LLaMA 1 is clearly shown for all classification tasks, especially binary and multilabel where LLaMA 2 has a  smaller number of irrelevant results and higher F1 score (see Figure~\ref{classtypeall}). 

   \begin{figure}[hbt!]
		    \begin{center}
		     \includegraphics[scale = 0.014]{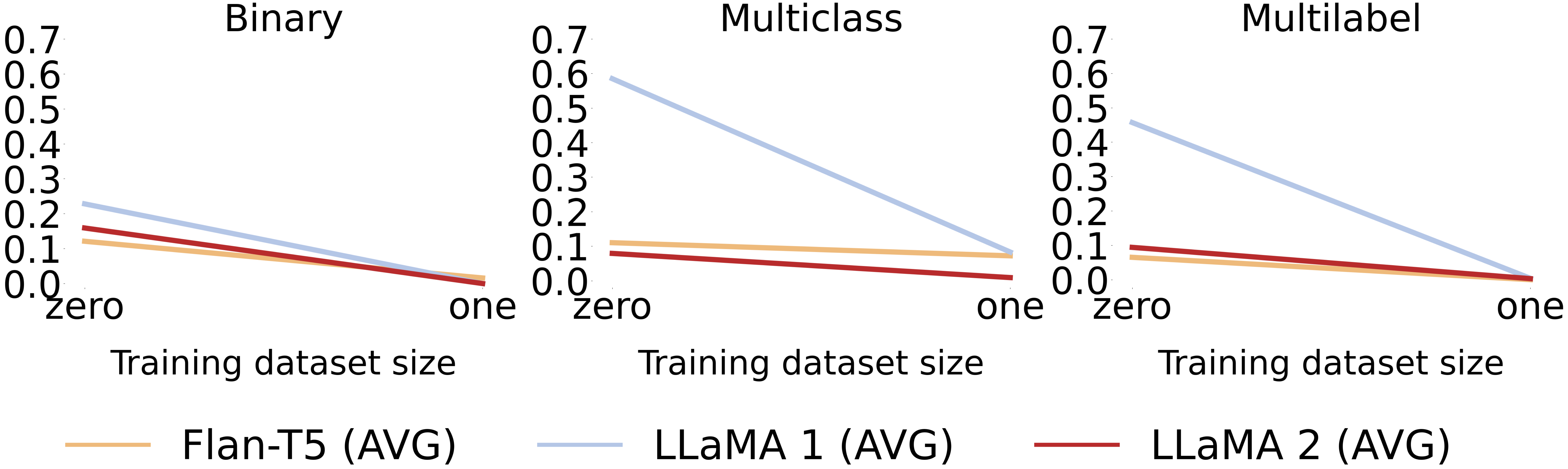}
		
        \caption{Wrong labels for prompting approaches per binary (left), multiclass (middle), and multilabel (right) classification where 'zero' refers to zero-shot learning and 'one' refers to one-shot learning.}\label{missinglabels}
		    \end{center}
    \end{figure}

 Regarding fine-tuning approaches, results in Figure~\ref{classtypeall} show a clear dominance of RoBERTa-large in one-shot setting for all classification types. 
 When fine-tuning is performed using the entire dataset, T5 outperforms the rest of the models for binary classification with micro-F1 = 0.672 versus RoBERTa-large with micro-F1 = 0.607. However, for multiclass and multilabeling tasks, the performance of T5 decreases and the model is outperformed by both RoBERTa-base and RoBERTa-large. For instance, for multiclass problems RoBERTa-large achieves micro-F1 of 0.726 versus micro-F1 for T5 = 0.700. For multilabeling problems the performance gap between the models increases and RoBERTa-large has a micro-F1 = 0.788 versus T5 with micro-F1 = 0.718. These results suggest that fine-tuned masked language models are more suitable for complex classification tasks such as multiclass and multilabeling problems when the number of labels is higher versus fine-tuning text-to-text models such as T5. 

A comparison between prompting and fine-tuning techniques for low resource settings suggests a better performance of prompting for binary and multiclass problems (see Figure~\ref{classtypeall}) where Flan-T5 and LLaMA 2 outperform fine-tuning models by a significant margin. For instance, Flan-T5 has micro-F1 = 0.553 versus micro-F1 for RoBERTa-large with micro-F1 = 0.485 for binary classification in one shot settings. The advantage of prompting in one shot settings becomes even more evident for multiclass problems where Flan-T5 achieves micro-F1 = 0.489 versus RoBERTa-large with micro-F1 = 0.162. However, for multilabeling problems, fine-tuning approaches outperform prompting methods with a difference in micro-F1 of 0.082 between the best fine-tuned model, RoBERTa-large, and the best prompting model, i.e. Flan-T5. It is worth noting that during one-shot training, all models have been provided the same training examples. 
However, further analyses are needed to identify most efficient ways for representing multi-labeling problems as part of prompting techniques. 
        
Despite the better overall performance of prompting techniques in zero- and one- shot settings, these approaches lead to unsatisfactory performance when compared to fine-tuned masked language models on a larger training set. Further, the difference in the performance between the two techniques grows larger for more complex text classification tasks such as multiclass and multilabeling problems. For instance, for binary classification, the difference in performance in terms of micro-F1 between best performing prompting and fine-tuning technique is 0.119 while for multiclass the difference in performance is 0.240. This shows that large autoregressive text generation models coupled with few shot learning techniques still have room for improvement when it comes to text classification. Fine-tuned masked language models, despite being smaller, lead to better performance for text classification versus LLMs in ICL settings.

\subsection{Trends across datasets and models}\label{comparison_bestperforming}
    
\begin{table*}[htbp]

	\centering
	\setlength{\tabcolsep}{6.0pt}
	\scalebox{0.68}{
		\begin{tabular}{|l|l||c|c|c|c|c|c|c|c|}\hline
		\multirow{2}{*}{\textbf{Dataset}}&\multirow{2}{*}{\textbf{Model}}&\multicolumn{3}{|c|}{\textbf{zero shot}}&\multicolumn{3}{|c|}{\textbf{one shot}}&\multicolumn{2}{|c|}{\textbf{all}}\\\cline{3-10}
		&&\textbf{micro F1}&\textbf{macro F1}&\textbf{wrong labs}&\textbf{micro F1}&\textbf{macro F1}&\textbf{wrong labs}&\textbf{micro F1}&\textbf{macro F1}\\\hline\hline
	  \multirow{4}{*}{irony}
        & RoBERTa&--&--&--&.459 ($\pm$.005)&.459 ($\pm$.005)&--&.508&.508\\\cline{2-10}
        &T5&---&---&---&.455($\pm$ .021)&.455($\pm$ .021)&.589&\textbf{.688}&\textbf{.688}\\\cline{2-10}
		&FlanT5&.428&.428&\underline{.049}&\textbf{.491($\pm$ .034)}&\textbf{.491($\pm$.034)}&.009&--&--\\\cline{2-10}
        &LLaMA&\textbf{.499}&\textbf{.499}&.214&.443 ($\pm$.003)&.443 ($\pm$.003)&\underline{.000}&--&--\\\cline{2-10}
        &GPT 3.5{*}&.727&.727&.000&--&--&--&--&--\\\hline\hline
          \multirow{4}{*}{offense}
        & RoBERTa&--&--&--&.550 ($\pm$.143)&.550 ($\pm$.143)&--&.705&.705\\\cline{2-10}
        &T5&--&--&--&.462 ($\pm$.001)&.462 ($\pm$.001)&.864&\textbf{.709}&\textbf{.709}\\\cline{2-10}
		&FlanT5&\textbf{.429}&\textbf{.429}&\underline{.269}&\textbf{.558($\pm$.019)}&\textbf{.558($\pm$.019)}&.003&--&--\\\cline{2-10}
        &LLaMA&.419&.419&.227&.347 ($\pm$.026)&.347 ($\pm$.026)&\underline{.001}&--&--\\\cline{2-10}
        &GPT 3.5{*}&.635&.635&.000&--&--&--&--&--\\\hline\hline
        \multirow{4}{*}{hate}
        & RoBERTa&--&--&--&.445 ($\pm$.118)&.445 ($\pm$.118)&--&.607&.607\\\cline{2-10}
        &T5&---&---&--&.386 ($\pm$.312)&.386 ($\pm$.312)&.732&\textbf{.619}&\textbf{.619}\\\cline{2-10}
		&FlanT5&\textbf{.634}&\textbf{.634}&\underline{.004}&\textbf{.611 ($\pm$.006)}&\textbf{.611 ($\pm$.006)}&.005&--&-\\\cline{2-10}
        &LLaMA&.539&.539&\underline{.004}&.514 ($\pm$.111)&.514 ($\pm$.111)&\underline{.000}&--&--\\\hline\hline
      
         \multirow{4}{*}{emoji}
        & RoBERTa&---&---&--&.047 ($\pm$.009)&.005 ($\pm$.001)&--&\textbf{.366}&\textbf{.317}\\\cline{2-10}
        &T5&---&---&---&.000 ($\pm$.000)&.000 ($\pm$.000)&.100&.259&\textbf{.317}\\\cline{2-10}
		&FlanT5&\textbf{.059}&\textbf{.042}&\underline{.036}&\textbf{.114($\pm$.021)}&\textbf{.082($\pm$.007)}&.006&--&--\\\cline{2-10}
        &LLaMA&.060&.041&.091&.033($\pm$.036)&.020($\pm$.017)&\underline{.001}&--&--\\\hline\hline
         \multirow{4}{*}{sentiment}
        & RoBERTa&---&---&--&.449 ($\pm$.108)&.271 ($\pm$.008)&--&\textbf{.714}&\textbf{.714}\\\cline{2-10}
        &T5&---&--&--&.312 ($\pm$.121)&.272 ($\pm$.078)&.563&.708&.709\\\cline{2-10}
		&FlanT5&\textbf{.459}&\textbf{.402}&.109&.417 ($\pm$.004)&.381 ($\pm$.007)&\underline{.000}&--&--\\\cline{2-10}
        &LLaMA&.369&.334&\underline{.027}&\textbf{.482 ($\pm$.037)}&\textbf{.402 ($\pm$.143)}&\underline{.000}&--&--\\\hline\hline
        \multirow{4}{*}{BBC}
        & RoBERTa&---&---&--&.217 ($\pm$.027)&.112 ($\pm$.029)&--&\textbf{.989}&\textbf{.989}\\\cline{2-10}
        &T5&---&---&--&.001 ($\pm$.001)&.001 ($\pm$.001)&.999&.977&.977\\\cline{2-10}
		&FlanT5&\textbf{.922}&\textbf{.867}&.096&\textbf{.939($\pm$.008)}&\textbf{.936($\pm$.009)}&.038&--&--\\\cline{2-10}
        &LLaMA&.498&.439&\underline{.021}&.849 ($\pm$.098)&.843 ($\pm$.081)&\underline{.004}&--&--\\\cline{2-10}
        &GPT 3.5{*}&.912&.913&.000&--&--&--&--&--\\\hline\hline
        \multirow{4}{*}{Reuters}
        & RoBERTa&---&---&--&.154 ($\pm$.111)&.054 ($\pm$.021)&--&\textbf{.939 }&\textbf{.869 }\\\cline{2-10}
        &T5&---&---&--&.010 ($\pm$.034)&.010 ($\pm$.067)&.990&.929&.833\\\cline{2-10}
		&FlanT5&\textbf{.321}&\textbf{.334}&.334&.467 ($\pm$.023)&\textbf{.504 ($\pm$.032)}&.017&--&--\\\cline{2-10}
        &LLaMA&.212&.168&\underline{.006}&\textbf{.528 ($\pm$.076)}&.304 ($\pm$.145)&\underline{.006}&--&--\\\cline{2-10}
        &GPT 3.5{*}&.852&.718&.000&--&--&--&--&--\\\hline\hline
        \multirow{4}{*}{20 News(all)}
        & RoBERTa&---&---&--&.190 ($\pm$.028)&.101 ($\pm$.019)&--&.859&.853\\\cline{2-10}
        &T5&---&---&--&.000 ($\pm$.000)&.000 ($\pm$.000)&.999&\textbf{.861}&\textbf{.854}\\\cline{2-10}	
        &FlanT5&\textbf{.564}&\textbf{.520}&\textbf{.001}&\textbf{.684 ($\pm$.008)}&\textbf{.654 ($\pm$.007)}&.057&--&--\\\cline{2-10}
        &LLaMA&.324&.272&.094&.368 ($\pm$.034)&.300($\pm$.079)&\underline{.001}&--&--\\\hline\hline
         \multirow{4}{*}{20 News(subcl)}
        & RoBERTa&---&---&--&.055 ($\pm$.013)&.015 ($\pm$.003)&--&\textbf{.741}&\textbf{.728}\\\cline{2-10}
        &T5&---&---&--&.000 ($\pm$.000)&.000 ($\pm$.000)&.990&.717&.693\\\cline{2-10}
		&FlanT5&\textbf{.510}&\textbf{.507}&\underline{.000}&\textbf{.512 ($\pm$.013)}&\textbf{.501 ($\pm$.013)}&\underline{.011}&--&--\\\cline{2-10}
        &LLaMA&.185&.194&.167&.376 ($\pm$.015)&.342 ($\pm$.014)&.020&--&--\\\hline\hline
        \multirow{4}{*}{Ohsumed}
        & RoBERTa&---&---&--&.025 ($\pm$.019)&.002 ($\pm$.004)&--&\textbf{.476}&\textbf{.415}\\\cline{2-10}
        &T5&---&---&--&.002 ($\pm$.001)&.002 ($\pm$.001)&.958&.452&.362\\\cline{2-10}
		&FlanT5&\textbf{.306}&\textbf{.283}&.194&\textbf{.288 ($\pm$.003)}&\textbf{.241 ($\pm$.001)}&.375&--&--\\\cline{2-10}
        &LLaMA&.151&.099&\underline{.154}&.180 ($\pm$.110)&.162 ($\pm$.110)&\underline{.036}&--&--\\\hline\hline
        \multirow{4}{*}{Toxic}
        & RoBERTa&---&---&--&.671 ($\pm$.005)&.550 ($\pm$.003)&--&.899&\textbf{.782}\\\cline{2-10}
        &T5&---&---&--&.020 ($\pm$.001)&.010 ($\pm$.011)&.989&\textbf{.913}&.661\\\cline{2-10}
		&FlanT5&\textbf{.629}&\textbf{.380}&\underline{.140}&\textbf{.710 ($\pm$.066)}&\textbf{.262 ($\pm$.014)}&\underline{.003}&--&--\\\cline{2-10}
        &LLaMA&.331&.142&.211&.005 ($\pm$.079)&.002 ($\pm$.077)&.004&--&--\\\hline\hline
        \multirow{4}{*}{Legal}
        & RoBERTa&---&---&--&.429 ($\pm$.030)&.285 ($\pm$.030)&--&.965&.601\\\cline{2-10}
        &T5&---&---&--&.500 ($\pm$.037)&.125 ($\pm$.042)&.970&\textbf{.982}&\textbf{.612}\\\cline{2-10}
		&FlanT5&\textbf{.251}&\textbf{.233}&\underline{.000}&\textbf{.351 ($\pm$.047)}&\textbf{.352 ($\pm$.028)}&\underline{.000}&--&--\\\cline{2-10}
        &LLaMA&.224&.167&.069&.269 ($\pm$.091)&.232 ($\pm$.175)&.005&--&--\\\hline\hline
        \multirow{4}{*}{Cancer}
        & RoBERTa&---&---&--&.309 ($\pm$.003)&.290 ($\pm$.002)&--&\textbf{.524}&\textbf{.414}\\\cline{2-10}
        &T5&---&---&--&.000 ($\pm$.000)&.000 ($\pm$.000)&.000&.344&.157\\\cline{2-10}
		&FlanT5&\textbf{.296}&\textbf{.286}&.246&\textbf{.361 ($\pm$.027)}&\textbf{.319 ($\pm$.017)}&\underline{.000}&--&--\\\cline{2-10}
        &LLaMA&.249&.178&\underline{.141}&.168 ($\pm$.131)&.104 ($\pm$.098)&.004&---&--\\\hline\hline
         \multirow{4}{*}{PCL}
        & RoBERTa&---&--&--&.555 ($\pm$.004)&.518 ($\pm$.006)&--&\textbf{.719}&\textbf{.592}\\\cline{2-10}
        &T5&---&---&--&.001 ($\pm$.001)&.001 ($\pm$.001)&.999&.654&.525\\\cline{2-10}
		&FlanT5&.224&.124&\underline{.000}&.224 ($\pm$.008)&.141 ($\pm$.112)&.001&--&--\\\cline{2-10}
        &LLaMA&\textbf{.392}&\textbf{.303}&.050&\textbf{.287 ($\pm$.095)}&\textbf{.159 ($\pm$.114)}&\underline{.000}&--&--\\\cline{2-10}
         &GPT 3.5{*}&.207&.117&.000&--&--&--&--&--\\\hline\hline
         \multirow{4}{*}{Safeguard(all)}
        & RoBERTa&---&---&--&.601 ($\pm$.011)&.589 ($\pm$.011)&--&\textbf{.905}&\textbf{.895}\\\cline{2-10}
        &T5&---&---&--&.000 ($\pm$.000)&.000 ($\pm$.000)&.000&.756&.725\\\cline{2-10}	
        &FlanT5&\textbf{.347}&\textbf{.326}&\underline{.000}&\textbf{.392 ($\pm$.007)}&\textbf{.360 ($\pm$.003)}&\underline{.000}&--&--\\\cline{2-10}
        &LLaMA&.291&.233&.041&.286 ($\pm$.007)&.197 ($\pm$.003)&.001&--&--\\\cline{2-10}
        &GPT 3.5{*}&.369&.340&.000&--&--&--&--&--\\\hline\hline
         \multirow{4}{*}{Safeguard(subcl)}
        & RoBERTa&---&---&--&.247 ($\pm$.105)&.201 ($\pm$.102)&--&\textbf{.718}&.515\\\cline{2-10}
        &T5&---&---&--&.010 ($\pm$.002)&.020 ($\pm$.002)&.969&.657&\textbf{.516}\\\cline{2-10}
       
         &FlanT5&\textbf{.275}&\textbf{.253}&\underline{.000}&\textbf{.281 ($\pm$.012)}&\textbf{.265 ($\pm$.009)}&\underline{.000}&--&--\\\cline{2-10}
        &LLaMA&.195&.176&.051&.237 ($\pm$.049)&.231 ($\pm$.089)&.006&--&--\\\cline{2-10}
          &GPT 3.5{*}&.359&.366&.012&--&--&--&--&--\\\hline\hline

        \end{tabular}
		}
	\caption{Micro-F1 and Macro-F1 results per dataset for RoBERTa (large), fine-tuned T5, Flan-T5, LLaMA 2, and GPT 3.5-Turbo. The ratio of wrongly-formatted outputs is included in the wrong labels (labs) column.The results for Flan-T5 and LLaMA 2 are based on averaged results across all prompts.}\label{tab:perdataset}
	\end{table*} 
 
 \begin{table*}[htbp]

	\centering
	\setlength{\tabcolsep}{6.0pt}
	\scalebox{0.68}{
		\begin{tabular}{|l|l||c|c|c|c|c|c|c|c|}\hline
		\multirow{2}{*}{\textbf{Dataset}}&\multirow{2}{*}{\textbf{Model}}&\multicolumn{3}{|c|}{\textbf{zero shot}}&\multicolumn{3}{|c|}{\textbf{one shot}}&\multicolumn{2}{|c|}{\textbf{all}}\\\cline{3-10}		
    &&\textbf{micro F1}&\textbf{macro F1}&\textbf{wrong labs}&\textbf{micro F1}&\textbf{macro F1}&\textbf{wrong labs}&\textbf{micro F1}&\textbf{macro F1}\\\hline\hline
          \multirow{4}{*}{IMDB}  
        & RoBERTa&---&---&--&.436 ($\pm$.311)&.436 ($\pm$0.311)&--&\textbf{.955}&\textbf{.955}\\\cline{2-10}
        &T5&---&---&----&.751 ($\pm$.065)&.751 ($\pm$.065)&.711&.952&.952\\\cline{2-10}
		&FlanT5&\textbf{.948}&\textbf{.948}&\underline{.097}&\textbf{.900 ($\pm$.007)}&\textbf{.900 ($\pm$.007)}&.017&--&--\\\cline{2-10}
        &LLaMA&.628&.628&.219&.803 ($\pm$.012)&.803 ($\pm$.012)&\underline{.005}&--&--\\\hline\hline

        \multirow{4}{*}{AG News}
   
        & RoBERTa&---&---&--&.280 ($\pm$.022)&.111 ($\pm$.024)&--&.906&.884\\\cline{2-10}
        &T5&---&---&--&.010 ($\pm$.003)&.010 ($\pm$.003)&.990&\textbf{.907}&\textbf{.886}\\\cline{2-10}
		&FlanT5&\textbf{.819}&\textbf{.789}&\underline{.000}&\textbf{.813 ($\pm$.008)}&\textbf{.782($\pm$.009)}&\underline{.000}&--&--\\\cline{2-10}
        &LLaMA&.479&.463&.011&.787 ($\pm$.006)&.753 ($\pm$.005)&.003&--&--\\\hline\hline

        \end{tabular}
		}
	\caption{Micro- and Macro-F1 results for `AG News' and `IMDB' datasets for RoBERTa-large, fine-tuned T5 model, Flan-T5, LLaMA 2. The ratio of wrongly-formatted outputs is included in the wrong labels (labs) column. The results for Flan-T5 and LLaMA 2 are based on averaged results across all prompts.}\label{tab:perpretraineddataset}
	\end{table*}

Results presented in Table~\ref{tab:perdataset} confirm findings from Section~\ref{promptvsfinetune} showing a clear dominance of Flan-T5 over LLaMA for zero- and one-shot prompting for the majority of datasets. Exceptions are the 'irony', 'sentiment', and 'PCL' datasets where LLaMA performs better for either zero or one shot setting, or both. For some datasets such as `hate', prompting models give better performance in zero- shot than one-shot setting. However, models still improve performance for these datasets in terms of number of wrong labels. Further, the choice of one shot training instances can influence the performance of models in few-shot learning. For the purposes of this analysis we have selected the one shot examples randomly. Analysing the impact of the training examples in few-shot learning can be a future research direction which we leave for future work. 

In contrast to the prompting approaches, results for the fine-tuned models do not show a clear dominance of either RoBERTa or T5. T5 shows a better performance for the majority of the binary classification tasks (those associated with Twitter datasets) as well as the datasets `AG news', `20 News' (top 6 classes)', and the `legal' domain. The two models attain a similar macro-F1 for the emoji prediction and safeguarding reports datasets.

\begin{figure}[hbt!]
		    \begin{center}
		     \includegraphics[scale = 0.012]{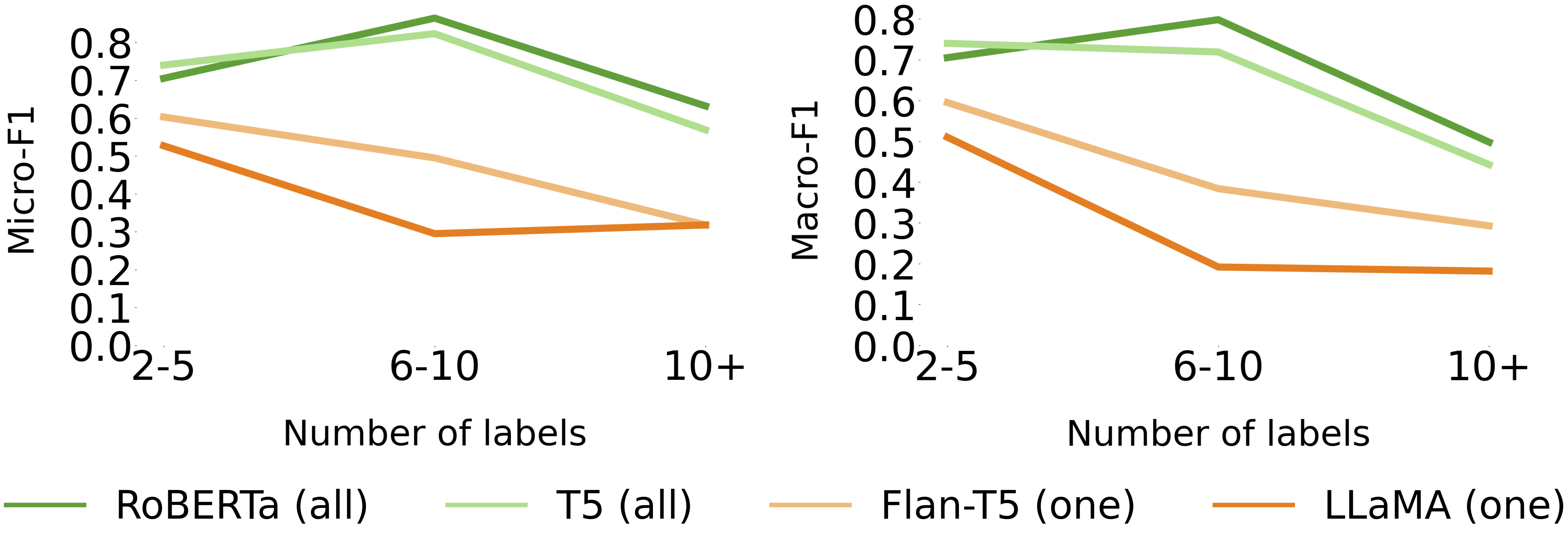}
      \caption{Averaged Micro-F1 and Macro-F1 results based on number of classification labels: `RoBERTa (all)' and `T5 (all)' refer to models fine-tuned on the entire training set, `Flan-T5 (one)' and `LLaMA (one)' refer to one-shot prompting.}\label{numlabels}
		    \end{center}
        \end{figure}

\noindent \textbf{Impact of the number of labels.}  
    Analysis into the effect of the number of classification labels in the performance shows an interesting trend with the fine-tuned models (RoBERTa and T5) performing slightly better for classification tasks with 6 to 9 labels than classification with less labels (see Figure \ref{numlabels}). For RoBERTa this trend occurs for both micro-F1 and macro-F1 while for T5 it appears only for micro-F1. This can be attributed to the nature of the binary classification tasks ('irony', 'offense', 'hate') which express human emotions and represent the Twitter domain. This suggests that the models find it more challenging to categorise such texts versus more categorical-based datasets such as news and articles which are part of the datasets with 6 to 9 labels. In contrast, the performance of both prompting approaches decreases as the number of labels for the classification task increases.

        \noindent \textbf{Datasets used for pre-training.} As mentioned earlier in the section, we analyse the performance of models for the ‘IMDB reviews’ and ‘AG News’ datasets separately as they are used in the fine-tuning of the Flan-T5 model. For these datasets (see Table~\ref{tab:perpretraineddataset}) Flan-T5 performance significantly improves achieving micro- and macro-F1 results comparable to fine-tuning models on the entire dataset. For instance, for the IMDB dataset, the difference in macro-F1 between Flan-T5 and RoBERTa is 0.007 while for the AG news the difference in macro-F1 is 0.027.  In contrast, the performance gap for the rest of the datasets between Flan-T5 and the best performing fine-tuning model is on average around 0.250 in micro-F1. This shows the significant impact that data contamination may have in the final results. However, a careful data contamination analysis becomes harder on large models for which training data is not available, and especially for closed models. 
	
     \noindent \textbf{GPT Analysis.} Table~\ref{tab:perdataset} presents zero-shot prompting results for the GPT 3.5-Turbo model for the following datasets: `irony', `offense', `bbc', `reuters', `pcl', and `safeguard'. We have used the class-based prompt for prompting with GPT 3.5 because it has shown to lead to the higher overall performance  for Flan-T5 and LLaMA. Results show a clear advantage of the GPT-based model over Flan-T5 and LLaMA achieving on average 0.350 higher micro- and macro-F1 across the majority of the datasets, except for the `PCL' dataset. Additionally, results achieved with zero-shot learning with GPT 3.5-Turbo outperform fine-tuned models on the entire dataset for the `irony' dataset. However, for the rest of the datasets the model is still outperformed by fine-tuning approaches confirming the lack of generalisation abilities of few-shot learning techniques and text generation models for text classification.

    \section{Conclusions}\label{conclusions}
This paper presents a large-scale study on how prompt-based LLMs in zero- and one- shot settings compare to smaller but fine-tuned language models for text classification. The evaluation spans across 16 datasets covering binary, multiclass, and multilabel problems. In particular, we compared three different types of models, i.e., linear models such as FastText, masked language models (RoBERTa), and text generation models tested in ICL settings (T5, Flan-T5, and LLaMA, as well as GPT 3.5-Turbo). Analyses on prompting techniques showed a clear advantage of the Flan-T5 model over LLaMA 1 and LLaMA 2 regardless of the prompt used for both zero- and one-shot settings. This shows that smaller but instruction-tuned models have better generalisation abilities for text classification than larger text generation models. Further, our analysis showed that results from zero- and few-shot learning LLMs are considerably lower in comparison to smaller models fine-tuned on the entire training set. This highlights the need for training data, even in the age of LLMs, and that fine-tuning smaller and more efficient language models can still outperform in-context learning methods of larger text generation models.

\section{Acknowledgements}

Aleksandra Edwards and Jose Camacho-Collados are supported by a UKRI Future Leaders Fellowship.

The safeguarding documents used for performing analysis in the paper have been collected in collaboration with the Wales Safeguarding Repository (WSR) project, funded by the National Independent Safeguarding Board (NISB), the Crime and Security Research Institute at Cardiff University (CSRI), and the School of Social Sciences at Cardiff University (SOCSI). We would like to thank the WSR team for their support.

\section{Limitations}
The main limitation of this research is the lack of experiments on fine-tuning Flan-T5 and LLaMA models as well as the lack of further analysis with larger text generation models such as LLaMA with 13 and 17 billion parameters. Moreover, the paper presents a study for zero- and one-shot prompting. As future work, we plan to extend analysis to understand how the number of training instances affect the performance of in-context learning approaches. Further, considering the sensitivity of in-context learning approaches to the given instructions, it would be beneficial to perform further analysis on a larger more diverse set of prompts. Finally, the paper presents results for a single high resource language (English). Experiments for other languages (especially low-resource) could show a different tendency.

\end{spacing}

\bibliography{custom}

\appendix

\section{Appendix}
In Section~\ref{promptvsfinetuneMacro} we present a comparison between prompting and fine-tuning techniques based on Macro-F1.
In Section~\ref{promptanalysis}, we present the prompts we used for performing analysis with zero- and one- shot in-context learning with Flan-T5 and LLaMA 1 and LLaMA 2.  
\subsection{Prompting versus Fine-tuning: Macro Results}\label{promptvsfinetuneMacro}
\begin{figure}
      \begin{center}
		     \includegraphics[scale = 0.03]{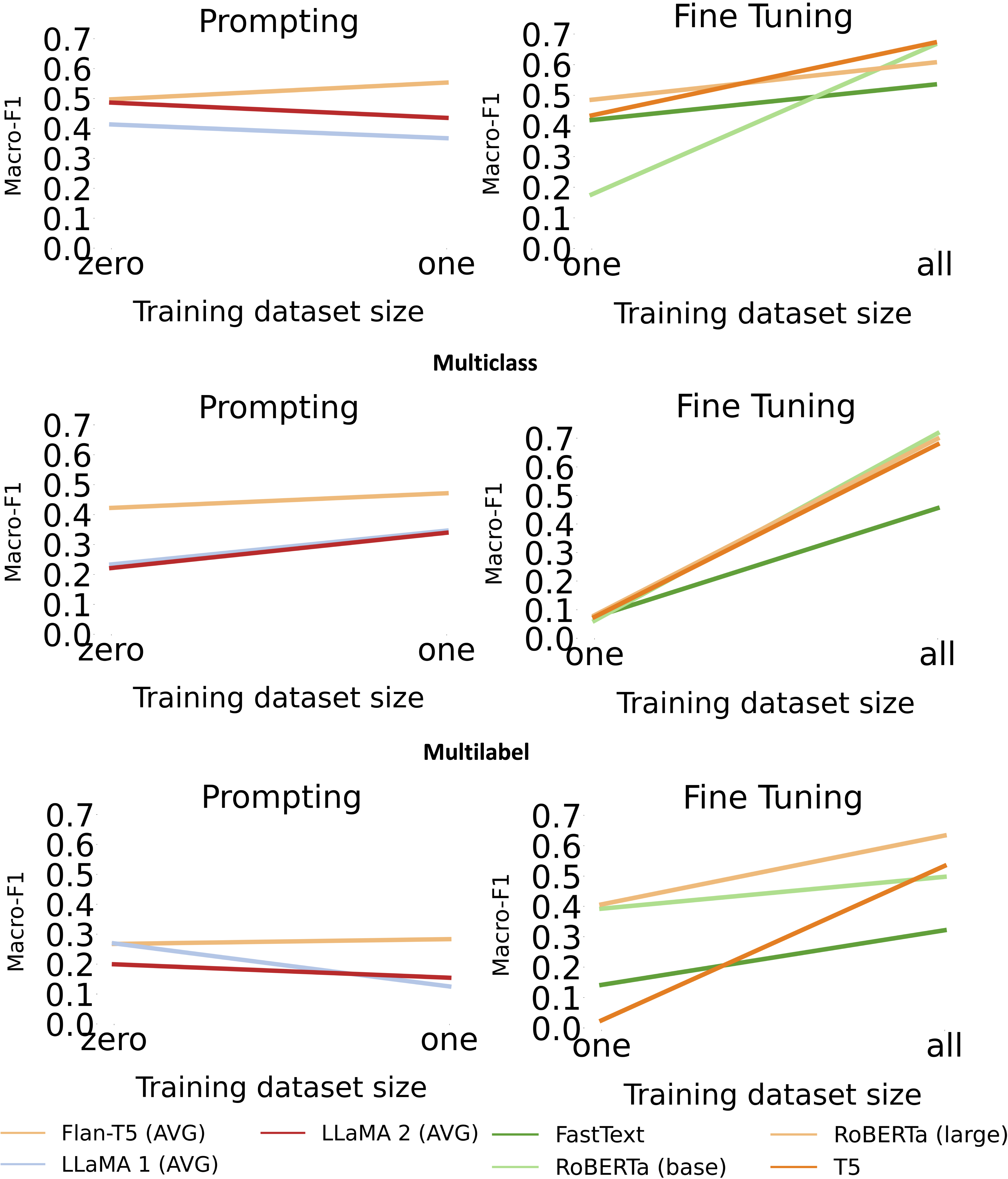}
		  
        \caption{Comparison between prompting (left) and fine-tuning (right) approaches per text classification type  where 'AVG' refers to averaged results across all prompt types per model. In 'Prompting', 'zero' and 'one' refer to zero- and one- shot prompt-based learning techniques, in 'Fine Tuning', 'one' refers to fine-tuning the models with one training instance per label and 'all' refers to fine-tuning using the entire dataset.}\label{classtypemacro}
		    \end{center}
        \end{figure}

Figure~\ref{classtypemacro} shows the Macro-F1 results comparing prompting and fine-tuning techniques. Results show similar trends to those observed based on Micro-F1, presented in Section~\ref{promptvsfinetune}.

   \subsection{Prompts}\label{promptanalysis}
   In Table~\ref{tab:allprompts} we have listed all `domain' prompts we used per dataset. The `task' and the `generic' prompts are the same for all datasets and are presented in Table~\ref{tab:prompts} in Section~\ref{prompts}. 
     \begin{table}[!hbt]
	\centering
	\scalebox{0.6}{
	\begin{tabular}{l|p{8.0cm}}\hline
		\textbf{Dataset}&\textbf{Domain Prompt}\\\hline
		irony&Is the Tweet classified as irony or non-irony?\\\hline
        offense&Is the Tweet classified as offensive or non-offensive?\\\hline
        hate&Is the Tweet classified as hate or non-hate?\\\hline
        emoji&Which of the given emojis best describe the given Tweet?The emojis are:\\\hline
        sentiment&Is the Tweet positive, negative, or neutral?\\\hline
        BBC&Classify the news into one of the following topics:\\\hline
        Reuters&Classify the news into one of the following topics:\\\hline
        20 News&Classify the newsgroup into one of the following topics:\\\hline
        Ohsumed&Select the medical conditions that this article is about. The options are:\\\hline
        Toxic&Which of the given toxic topics best describe the given comment? Choose one or more from the following topics:\\\hline
        Legal&Which of the given legal topics best describe the given legislation document?Choose one or more from the following topics:\\\hline
        Cancer&Which hallmarks of cancer are present in the text? Choose one or more from the following options\\\hline
        PCL&Which of the given topics best describe the patronising comment. Choose one or more from the following topics:\\\hline
        Safeguard&Which of the given themes best describe the sentence? Choose one or more from the following themes:\\\hline
        IMDB&Is the movie review positive or negative?\\\hline
        AG News&Select the topic that the given article is about.The topics are:\\\hline

	\end{tabular}
	}
	\caption{A list of all domain-based prompts used per dataset.}\label{tab:allprompts}
	\end{table}

\end{document}